\documentclass[runningheads,a4pasper]{llncs}

\usepackage{amssymb}
\usepackage{graphicx}
\usepackage{url}
\urldef{\mailsa}\path|{ssisovic, smarti, amestrovic}@uniri.hr|   
\newcommand{\keywords}[1]{\par\addvspace\baselineskip
\noindent\keywordname\enspace\ignorespaces#1}

\begin{document}

\mainmatter

\title{Toward Network-based Keyword Extraction from Multitopic Web Documents}

\titlerunning{Toward Network-based Keyword Extraction from Multitopic Documents}

\author{Sabina \v{S}i\v{s}ovi\'c, Sanda Martin\v{c}i\'c-Ip\v{s}i\'c, Ana Me\v{s}trovi\'c}

\authorrunning{Sabina \v{S}i\v{s}ovi\'c, Sanda Martin\v{c}i\'c-Ip\v{s}i\'c, Ana Me\v{s}trovi\'c}

\institute{Department of Informatics,\\
University of Rijeka,\\
Radmile Matej\v{c}i\'c 2, 51000 Rijeka, Croatia \\
\mailsa\\
}

\maketitle

\begin{abstract}
In this paper we analyse  the selectivity measure calculated from the complex network in the task of the automatic keyword extraction. Texts, collected from different web sources (portals, forums), are represented as directed and weighted co-occurrence complex networks of words. Words are nodes and links are established between two nodes if they are directly co-occurring within the sentence. We test different centrality measures for ranking nodes - keyword candidates. The promising results are achieved using the selectivity measure. Then we propose an approach which enables extracting word pairs according to the values of the in/out selectivity and weight measures combined with filtering.  

\end{abstract}

\section{Introduction}

Keyword extraction is an important task in the domain of the Semantic Web development. It is a problem of automatic identification of the important terms or phrases in text documents. It may have numerous applications: information retrieval, automatic indexing, text summarization, semantic description and classification, etc. In the case of web documents it is a very demanding task: it requires extraction of keywords from web pages that are typically noisy, overburden with information irrelevant to the main topic (navigational information, comments, future announces, etc.) and they usually contain several topics \cite{grineva2009extracting}. Therefore, in keyword extraction from web pages we are dealing with noisy and multitopic datasets.

Various approaches have been proposed for keywords and keyphrases identification (extraction) task. There are two main classes of approaches: supervised and unsupervised. Supervised approaches are based on using machine learning techniques on the manually annotated data \cite{witten99,turney2000}. Therefore supervised approaches are time consuming and expensive. Unsupervised approaches may include clustering \cite{liu}, language modelling \cite{tomokiyo} and graph-based approaches. Unsupervised approaches may also require different sets of external data, however these approaches are not depended on manual annotation. These approaches are more robust, but usually less precise \cite{boudin2013comparison}.

A class of graph-based keyword extraction algorithms overcome some of these problems. In graph-based or network-based approaches the text is represented as a network in a way that words are represented as nodes and links are established between two nodes if they are co-occurring within the sentence. The main idea is to use different centrality measures for ranking nodes in the network. Nodes with the highest rank represent words that are candidates for keywords and keyphrases. In \cite{lahiri2014keyword} an exhaustive overview of network centrality measures usage in the keyword identification task is given. 

One of the probably most influential graph-based approaches is the TextRank ranking model introduced by Mihalcea and Tarau in \cite{mihalcea2004textrank}. TextRank is a modification of PageRank algorithm and the basic idea of this ranking technique is to determine the importance of a node according to the importance of its neighbours, using global information recursively drawn from the entire network. However, some recent researches have shown that even simpler centrality measures can give satisfactory results. Boudin \cite{boudin2013comparison}  and Lahiri et al. \cite{lahiri2014keyword}  compare different centrality measures for keyword extraction task. Litvak and Last \cite{litvak2008graph} compare supervised and unsupervised approach for keywords identification in the task of extractive summarization.

We have already experimented with graph-based approaches for Croatian texts representation. In \cite{mestrovicDictionary,mestrovicCDUD} we described graph-based word extraction and representation from the Croatian dictionary. We used lattice to represent different semantic relations (partial semantic overlapping, more specific, etc.) between words from the dictionary. 
In \cite{margan2014shuffled,margan2014shuffled2,sisovic2014blogs} we described and analysed network-based represenattion of Croatian texts.
In \cite{margan2014shuffled2} our results showed that in and out selectivity values from shuffled texts are constantly below selectivity values calculated from normal texts. It seems that selectivity measure is able to capture typical word phrases and collocations which are lost during the shuffling procedure. The same holds for English where Masucci and Rodgers \cite{masucci2009differences} found that selectivity somehow captures the specialized local structures in nodes’ neighborhood and forms of the morphological structures in text. According to these results, we expected that node selectivity may be potentially important for the text categories differentiation and include it in the set of standard network measures. In \cite{sisovic2014blogs} we show that the node selectivity measure can capture the structural differences between two genres of text.

This was the motivation for further exploration of selectivity for keyword extraction task from Croatian multitopic web documents. We propose an in/out selectivity based approach combined with filtering to extract keyword candidates from the co-occurrence complex network of text. We design selectivity-based approach as unsupervised, data and domain independent. In its basic form, only the stopwords list is a prerequisite for applying stopwords-filter. As designed, it is a very simple and robust approach appropriate for extraction from large multitopic and noisy datasets.

In Section 2 we present measures for the network structure analysis. In Section 3 we describe datasets and the construction of co-occurrence networks from used text collection. In Section 4 are the results of keyword extraction, and in the final Section 5, we elaborate the obtained results and make conclusions regarding future work.

\section{The network measures}
This section describes basic network measures that are necessary for understanding our approach. More details about these measures can be found in \cite{newman2010networksbook,opsahl2010,masucci2009differences}.
In the network, $N$ is the number of nodes and  $K$ is the number of links. In weighted language networks every link connecting two nodes $i$ and $j$ has an associated weight $w_{ij}$ that is a positive integer number. 

The node degree $k_{i}$ is defined as the number of edges incident upon a node. The in degree and out degree $k_{i}^{in/out}$ of node $i$ is defined as the number of its in and out neighbours.

Degree centrality of the node $i$ is the degree of that node. It can be normalised by dividing it by the maximum possible degree $N-1$:

\begin{equation}
dc_{i} =  \frac{k_{i}}{N-1}.
\end{equation} 

Analogue, the in/out degree centralities are defined as in/out degree of a node:

\begin{equation}
dc_{i}^{(in/out)} =  \frac{k_{i}^{(in/out)}}{N-1}.
\end{equation} 

Closeness centrality is defined as the inverse of farness, i.e. the sum of the shortest distances between a node and all the other nodes. Let $d_{ij}$ be the shortest path between nodes $i$ and $j$. The normalised closeness centrality of a node $i$ is given by:

\begin{equation}
cc_{i} =  \frac{N-1}{\sum_{i \neq j}d_{ij}}.
\end{equation} 

Betweenness centrality quantifies the number of times a node acts as a bridge along the shortest path between two other nodes. Let $\sigma_{jk}$ be the number of shortest paths from node  $j$ to node $k$ and let $\sigma_{jk}(i)$ be the number of those paths that pass through the node $i$. The normalised betweenness centrality of a node $i$ is given by:

\begin{equation}
bc_{i} =  \frac{\sum_{i \neq j \neq k}\frac{\sigma_{jk}(i)}{\sigma_{jk}}}{(N-1)(N-2)}.
\end{equation} 

The strength of the node $i$ is a sum of weights of all links incident with the node $i$:

\begin{equation}
s_{i} =  \sum_{j}w_{ij}.
\end{equation}

All given measures are defined for directed networks, but language networks are weighted, therefore, the weights should be considered. In the directed network, the in/out strength $s_{i}^{in/out}$ of the node $i$ is defined as the number of its incoming and outgoing links, that is: 

\begin{equation}
s_{i}^{in/out} =  \sum_{j}w_{ji/ij}.
\end{equation} 

The selectivity measure is introduced in \cite{masucci2009differences}. It is actually an average strength of a node. For the node $i$ the selectivity is calculated as a fraction of the node weight and node degree:

\begin{equation}
e_{i} = \frac{s_{i}}{k_{i}}.
\end{equation}

In the directed network, the in/out selectivity of the node $i$ is defined as:

\begin{equation}
e_{i}^{in/out} = \frac{s_{i}^{in/out}}{k_{i}^{in/out}}.
\end{equation}

\section{Methodology}

\subsection{The construction of co-occurrence networks}
Dataset contains 4 collections of web documents written in Croatian language collected from different web sources (portals and forums on different daily topics).
The 4 different web sources: business portal Gospodarski list (GL), legislative portal Narodne novine (NN), news portal with forum Index.hr (IN), daily newspaper portal Slobodna Dalmacija (SD). 

The first step in networks construction was text preprocessing: “cleaning” special symbols and normalising Croatian diacritics (\v{c}, \'c, \v{z}, \v{s}, d\v{z}) and removing punctuation which does not mark the end of a sentence. Commonly, for Croatian which is highly flective Salvic language the lemmatisation and part-of-speech tagging should be performed, but we model our experiment without any explicit language knowledge.

For each dataset we constructed weighted and directed co-occurrence network. Nodes are words that are linked if they are direct neighbours in a sentence. The next step was creating the networks as weighted edgelists, which contain all the pairs of connected words and their weights (the number of connections between two same words). In the Table 1 there are number of words, number of nodes and number of edges per each dataset.
We used Python and the NetworkX software package developed for the construction, manipulation, and study of the structure, dynamics, and functions of complex networks \cite{hagberg2008exploring}.

\begin{table}

\caption{The number of words, number of nodes and number of edges for all 4 datasets}
\begin{center}

\begin{tabular}{|l|c|c|c|c|}
\hline Dataset&GL	& NN	& IN	& SD \\ \hline
Number of words &  199 417 &146 731	&118 548&	44 367\\ \hline 
Number of nodes $N$ &  27727	&13036&	15065&	9553\\ \hline 
Number of links $K$& 105171	&55661	&28972	&25155\\ \hline 
\end{tabular}
\end{center}
\end{table}

\subsection{The selectivity-based approach}
The goal of this experiment is to analyse the selectivity measure in the automatic keyword identification task. 
First, for each node three centrality measures: in/out degree centrality, closeness centrality and betwenness and selectivity. Then we rank all nodes (words) according to the values of each of these measures, obtaining top 10 keyword candidates automatically from the network.

In the second part of our experiment we compute in/out selectivity for all nodes in all 4 networks. The nodes are then ranked according to the highest in/out selectivity values. Then, for every node we detect neighbour nodes with the highest weight. For the in selectivity we isolate one neighbour node with the highest outgoing link weight. For the out selectivity we isolate one neighbour node with the highest ingoing link weight. The result of in/out selectivity extraction is a set of ranked words tuples.

The third part of our approach consider applying different filters on the in/out selectivity based word tuples. The first is the stopwords-filter: we filter out all tuples that contain stopwords. Stopwords are a list of the most common, short function words which do not carry strong semantic properties, but are needed for the syntax of language (pronouns, prepositions, conjunctions, abbreviations, interjections,...). The second is the high-weights-filter: from the in/out – selectivity based word tuples we chose only those tuples that have the same values for the selectivity and weight. The third filter is the combination of the first two filters.

\section{Results}

Initially, we analyse 4 networks constructed for each dataset. The top 10 ranked nodes with the highest values of the selectivity, in degree, out degree, closeness and betwenness measures for datasets IN, GL, SD and NN are shown in the Tables 2,3,4 and 5. It is obvious that top 10 ranked words according to the in/out degree centrality, closeness centrality and betwenness centrality are stopwords. It can be also noticed that centrality measures return almost identical top 10 stopwords. However, the selectivity measure ranked only open-class words: nouns, verbs and adjectives. We expect that among these highly ranked words are keyword candidates.

\begin{table}

\caption{Top ten words from the dataset IN ranked according to the selectivity, in/out degree, closeness and betwenness }
\begin{center}

\begin{tabular}{|l|c|c|c|c|c|}
\hline & selectivity & in degree & out degree & closeness & betweenness\\ \hline
1. & mladi\'cevi (joungsters) & i (and) &	i (and) & je (is) & i (in)\\ \hline 
2. & pomlatili (beaten) & u (in)& je (is) &  i (and) & je (is)   \\ \hline 
3. & seksualnog (sexual) & je (is)& u (in) & se (self) & u (in)   \\ \hline 
4. & policijom (police) & na (on)& na (on) & da (that) & na (on)   \\ \hline 
5. & uhi\'ceno (arrested)& da (that)& se (self) & su (are) & se (self)   \\ \hline 
6. & skandala (scandal) & za (for) & za (for) & to (it) & za (for)   \\ \hline 
7. & podnio (submitted) & se (self)& su (are)& a (but)& da (that)   \\ \hline 
8. & obo\v{z}avatelji (fans)& a (but)& da (that) & \'ce (will)& su (are)   \\ \hline 
9. & sata (hour)& su (are) & s (with)& samo (only) & a  (but) \\ \hline 
10. & Baskiji (Baskia)& s (with)& od (from)& ali (but)& s (with)   \\ \hline 

\end{tabular}
\end{center}
\end{table}

\begin{table}

\caption{Top ten words from the dataset GL ranked according to the selectivity, in/out degree, closeness and betwenness }
\begin{center}

\begin{tabular}{|l|c|c|c|c|c|}
\hline & selectivity & in degree & out degree & closeness & betweenness\\ \hline
1. & stupastih (cage) & i (and) & i (and) & i (and) & i (and) \\ \hline 
2. & populaciju (population) & u (in) & u (in) & se (self) & u (in)   \\ \hline 
3. & izdanje (issue) & na (on) & je (is) & je (is) & je (is)   \\ \hline 
4. & online (online)& je (is) &  se (self) & su (are) & na (on)   \\ \hline 
5. & webshop (webshop)& ili (or) & na (on) & a (but) & se (self)   \\ \hline 
6. & matrica (matrix) & a (but) & ili (or) & ili (or) & ili (or)   \\ \hline 
7. & pretplata (subscription) & se (self) & su (are)& to (it) & a (but)   \\ \hline 
8. & \v{c}asopis (journal) & za (for)& za (for) & bolesti (disease)& za (for)   \\ \hline 
9. & oglasi (ads) & od (from) & od (from) & da (that) & su (are)   \\ \hline 
10. & marketing (marketing)& su (are) & a (but) & biljke (plants)& od (from)   \\ \hline 

\end{tabular}
\end{center}
\end{table}

\begin{table}

\caption{Top ten words from the dataset SD ranked according to the selectivity, in/out degree, closeness and betwenness }
\begin{center}

\begin{tabular}{|l|c|c|c|c|c|}
\hline & selectivity & in degree & out degree & closeness & betweenness\\ \hline
1. & seronjo (bullshitter) & i (and) & i (and) & i (and) & i (and) \\ \hline 
2. & Splitu (Split) & u (in) & je (is) & je (is) & je (is)   \\ \hline 
3. & upi\v{s}ite (fill-in) & je (is) & u (in) & svibanj (May) & u (in)   \\ \hline 
4. & uredniku (editor) & komentar (comment) &  se (self) & se (self) & se (self)   \\ \hline 
5. & ekrana (monitor) & na (on) & svibanj (May) & ali (but) & na (on)   \\ \hline 
6. & crkvu (church) & se (self) & na (on) & a (but) & od (from)   \\ \hline 
7. & supetarski (Supetar) & za (for) & za (for) & će (will) & za (for)   \\ \hline 
8. & vijesti (news) & a (but) & da (that) & to (it) & a (but)   \\ \hline 
9. & zaradom (earning) & svibanj (May) & ne (ne) & još (more) & svibanj (May)   \\ \hline 
10. & Jovi\'c (Jovi\'c) & od (from) & a (but) & pa (so) & to (it)   \\ \hline 

\end{tabular}
\end{center}
\end{table}

\begin{table}

\caption{Top ten words from the dataset NN ranked according to the selectivity, in/out degree, closeness and betwenness }
\begin{center}

\begin{tabular}{|l|c|c|c|c|c|}
\hline & selectivity & in degree & out degree & closeness & betweenness\\ \hline
1. & novine (newspaper) & i (and) & i (and) & i (and) & i (and) \\ \hline 
2. & temelju (based on) & u (in) & u (in) & ili (or) & u (in)   \\ \hline 
3. & manjinu (minority) & za (for) & je (is) & je (is) & za (for)   \\ \hline 
4. & srpsku (Serbian) & na (on) &  za (for) & se (self) & ili (or)   \\ \hline 
5. & sladu (harmony) & ili (or) & se (self) & da (that) & na (on)   \\ \hline 
6. & snagu (strength) & iz (from) & ili (or) & usluga (service) & je (is)   \\ \hline 
7. & osiguranju (insurance) & te (and) & na (on) & zakona (law) & se (self)   \\ \hline 
8. & narodnim (national) & je (is) & o (on) & a (but) & o (on)   \\ \hline 
9. & novinama (newspaper) & se (self) & te (and) & skrbi (welfare) & te (and)   \\ \hline 
10. & kriza (crisis) & s (with) & \v{c}lanak (article) & HRT-a (HRT-a) & iz (form)   \\ \hline 

\end{tabular}
\end{center}
\end{table}

Furthermore, we analyse selectivity measure in details. Since texts are better represented as directed networks \cite{margan2013preliminary}, we analyse words with in selectivity and out selectivity measure separately. We extract words-tuple: the word before for in selectivity and the word after for out selectivity that has the highest value of the weight. In the Table 6 are shown ten highly ranked in/out selectivity based words-tuples together with the values of in/out–selectivity and weight. 

Hence, we extract the most frequent words-tuples which are possible collocations or phrases from the text. We expect that among these highly ranked words-tuples are keyword candidates. Regarding to the limited space, we show results only for the NN dataset, but other datasets raised similar results.

\begin{table}

\caption{Top ten highly ranked in/out selectivity based words-tuples for the NN dataset}
\begin{center}

\begin{tabular}{|c|l|c|c|l|c|c|}
\hline
& \multicolumn{3}{c|}{in selectivity}&\multicolumn{3}{c|}{out selectivity}\\
 & words tuple &$e^{in}$ & $w$
&words tuple&$e^{out}$&$w$\\ 
\hline
1. &narodne \textbf{novine} &326&326 &\textbf{srpsku} nacionalnu&222&222 \\ \hline 
2. &na \textbf{temelju}&317&317&\textbf{nacionalnu} pripadnost&183&1\\ \hline 
3. &nacionalnu \textbf{manjinu}&275&2&\textbf{ovjesne} jedrilice&159&159\\ \hline 
4. &za \textbf{srpsku}&222&222&\textbf{narodnim} novinama&129&129\\ \hline 
5. &u \textbf{skladu}&202&202&\textbf{narodne} jazz&111&1\\ \hline 
6. &na \textbf{snagu}&172&172&\textbf{manjinu} gradu&78&1\\ \hline 
7. &o \textbf{osiguranju}&134&43&\textbf{ovoga} sporazuma&72&1\\ \hline 
8. &u \textbf{narodnim}&129&129&\textbf{crvenog} kristala&72&3\\ \hline 
9. &narodnim \textbf{novinama}&129&129&\textbf{skladu} provjeriti&67&1\\ \hline 
10. &crvenog \textbf{kri\v{z}a}&99&2&\textbf{oru\v{z}ani}h sukoba&58&4\\ \hline 

\end{tabular}
\end{center}
\end{table}

In Table 6 there are words-tuples which contain stopwods, especially for the in selectivity based ranking.Therefore we use stopwords-filter defined in the previous section as shown in Table 7. Now we obtain more open class keyword candidates from highly ranked words-tuples.

\begin{table}

\caption{Top ten highly ranked in/out selectivity based words-tuples without stopwords for the NN dataset}
\begin{center}

\begin{tabular}{|c|l|l|c|c|l|l|c|c|}
\hline
& \multicolumn{3}{c|}{in selectivity}&\multicolumn{3}{c|}{out selectivity}\\
 & words tuple &$e^{in}$ & $w$
&words tuple&$e^{out}$&$w$\\ 
\hline
1. &narodne \textbf{novine}&326&326&\textbf{srpsku} nacionalnu&222&222 \\ \hline 
2. &nacionalnu \textbf{manjinu}&275&2&\textbf{nacionalnu} pripadnost&183&1 \\ \hline 
3. &narodnim \textbf{novinama}&129&129&\textbf{ovjesne} jedrilice&183&1\\ \hline 
4. &crvenoga \textbf{kri\v{z}a}&99&2&\textbf{narodnim} novinama&129&129\\ \hline 
5. &jedinicama \textbf{regionalne}&65&1&\textbf{narodne} jazz&111&1\\ \hline 
6. &nacionalne \textbf{manjine}&61&61&\textbf{manjinu} gradu&78&1\\ \hline 
7. &rizika \textbf{snaga}&57&1&\textbf{ovoga} sporazuma&72&1\\ \hline 
8. &medije \textbf{ubroj}&47&1&\textbf{crvenog} kristala&72&3\\ \hline 
9. &crveni \textbf{kri\v{z}}&42&42&\textbf{skladu} provjeriti&67&1\\ \hline 
10. &uopravni \textbf{spor}&41&41&\textbf{oru\v{z}anih} sukoba&58&4\\ \hline 

\end{tabular}
\end{center}
\end{table}

In the Table 8. there are 10 highly ranked words-tuples for the NN dataset with the high-weights-filter applied. Using this approach some new keyword candidates appear in the ranking results.

\begin{table}

\caption{Top ten highly ranked in/out selectivity based words-tuples with equal in/out selectivity and weight for the NN dataset }
\begin{center}

\begin{tabular}{|l|c|l|c|}
\hline
  \multicolumn{2}{|c|}{in selectivity} 
 & \multicolumn{2}{c|}{out selectivity}  \\
 \hline words tuple & $e^{in}$ = $w$
  & words tuple & $ e^{out}$=$w$\\ 
\hline 
na \textbf{temelju} (based on)&317&\textbf{ovjesne} jedrilice (hangh glider)&159\\ \hline 
za \textbf{srpsku} (for Serbian)&222&\textbf{narodnim} novinama (Nat. news.)&129\\ \hline 
u \textbf{skladu} (according to)&202&\textbf{sjedi\v{s}tem} u (headquarter in)&55\\ \hline 
na \textbf{snagu} (into effect)&172&\textbf{objavit} \'{c}e (will be bublished)&53 \\ \hline 
u \textbf{narodnim} (in national)&129&\textbf{republici} Hrvatskoj (Croatia)&52\\ \hline 
narodnim \textbf{novinama} (Nat. news.)&129&\textbf{albansku} nacionalnu (Albanian nat.)&52\\ \hline 
i \textbf{dopunama} (and amendments)&68&\textbf{republika} Hrvatska (Croatia)&49\\ \hline 
nacionalne \textbf{manjine} (nat. minority)&61&\textbf{oplemenjiva\v{c}kog} prava (noble law)&45\\ \hline 
sa \textbf{sjedi\v{s}tem} (with headquarter)&55&\textbf{madjarsku} nacionalnu (Hung. nat.)&40\\ \hline 

\end{tabular}
\end{center}
\end{table}

In the Table 9. there are 10 highly ranked words-tuples from the NN dataset with the both filters applied. According to our knowledge about the content of the dataset, these two filters derived the best results.

\begin{table}

\caption{Top ten highly ranked in/out selectivity based words-tuples with equal in/out
selectivity and weight without stopwords for the NN dataset}
\begin{center}

\begin{tabular}{|l|l|}
\hline \multicolumn{1}{|c|}{\textbf{in selectivity words tuple}} & \multicolumn{1}{|c|}{\textbf{out selectivity words tuple}} \\ \hline \hline
narodne \textbf{novine} (National newspaper)&\textbf{srpsku} nacionalnu (Serbian national) \\ \hline 
narodnim \textbf{novinama} (Nat. newspapers)&\textbf{ovjesne} jedrilice (hangh glider)\\ \hline 
nacionalne \textbf{manjine} (nat. minority)&\textbf{narodnim} novinama (Nat. newspapers)\\ \hline 
crveni \textbf{kri\v{z}} (red cross)&\textbf{republici} hrvatskoj (Republic of Croatia)\\ \hline 
upravni \textbf{spor} (administrative dispute)&\textbf{albansku} nacionalnu (Albanian national)\\ \hline 
ovjesnom \textbf{jedrilicom} (hangh glider)&\textbf{republika} hrvatska (Republic of Croatia)\\ \hline 
elektroni\v{c}ke \textbf{medije} (electronic media)&\textbf{oplemenjiva\v{c}kog} prava (noble law)\\ \hline 
nacionalnih \textbf{manjina} (national minority)&\textbf{madjarsku} nacionalnu (Hungarian nat.)\\ \hline 
domovinskog \textbf{rata} (Fatherland War)&\textbf{romsku} nacionalnu (Bohemian national)\\ \hline 
Ivan \textbf{Vrlji\'{c}} (Ivan Vrlji\'{c})&\textbf{nadzorni} odbor (supervisory board)\\ \hline

\end{tabular}
\end{center}
\end{table}

\section{Conclusion and discussion}

We analyse network-based keyword extraction from multitopic Croatian web documents using selectivity measure. We compare keyword candidate words rankings with selectivity and three network-based centrality measures (degree, closeness and betwenness). The selectivity measure gives better results because centrality-based rankings select only stopwords as the top 10 ranked words. Furthermore, we propose extracting the highly connected words-tuples with the highest in/out–selectivity values as the keyword candidates. Finally, we apply different filters (stopwords-filter, high-weights-filter) in order to keyword candidate list.

The first part of analysis can raise some considerations regarding the selectivity measure. The selectivity measure is important for the language networks especially because it can differentiate between two types of nodes with high strength values (which mean words with high frequencies). Nodes with high strength values and high degree values would have low selectivity values. These nodes are usually stopwords (conjunctions, prepositions, …). On the other side, nodes with high strength values and low degree values would have high selectivity values. These nodes are possible collocations, keyphrases and names that appear in the texts. It seems that selectivity is insensitive to stopwords (which are the most frequent words) and therefore can efficiently detect  semantically rich open class words from the network.

Furthermore, since we modelled multitopic datastes the keyword extraction task is even more demanding. From the results of this preliminary research it seems that the selectivity has a potential to extract keyword candidates without preprocessing (lemmatisation, POS tagging) from multitopic sources.

There are several drawbacks in this reported work: we did not perform the classical evaluation procedure because we did not have annotated data and we conducted analysis only on Croatian texts.

For the future work we plan to evaluate our results on different datasets in different languages. Furthermore, it seem promising to define an approach that can extract a sequence of three or four neighbouring words based on filtered words-tuples. We also plan to experiment with lemmatised texts.
Finally, in the future we will examine the effect of noise to the results obtained from multitopic sources.

\bibliographystyle{splncs}

\end{document}